\def\year{2020}\relax
\documentclass[letterpaper]{article} 
\usepackage{aaai20}  
\usepackage{times}  
\usepackage{helvet} 
\usepackage{courier}  
\usepackage[hyphens]{url}  
\usepackage{graphicx} 
\usepackage{graphicx}  
\makeatletter
\newcommand{\printfnsymbol}[1]{%
	\textsuperscript{\@fnsymbol{#1}}%
}
\makeatother

\usepackage[utf8]{inputenc} 
\usepackage{url}            
\usepackage{booktabs}       
\usepackage{amsfonts}       
\usepackage{nicefrac}       
\usepackage{microtype}      
\usepackage{graphicx}
\usepackage{algorithm}
\usepackage[noend]{algpseudocode}
\usepackage[para,online,flushleft]{threeparttable}
\usepackage{adjustbox}
\usepackage{enumitem}
\usepackage{makecell}

\usepackage{macro}

\renewcommand{\bb}{{\bf{b}}}
\newcommand{\bx}{{\bf{x}}}
\newcommand{\by}{{\bf{y}}}
\newcommand{\bu}{{\bf{u}}}

\newcommand{\bW}{{\bf{W}}}
\newcommand{\R}{{\mathbb{R}}}

\title{Dropout Prediction over Weeks in MOOCs via \\ Interpretable Multi-Layer Representation Learning}
\newcommand\equalcontrib{\thanks{Equal contribution}}
\author{
	\Large \textbf{Byungsoo Jeon\textsuperscript{\rm 1}\equalcontrib, Namyong Park\textsuperscript{\rm 1}\printfnsymbol{1}, Seojin Bang\textsuperscript{\rm 1}\printfnsymbol{1}}\\ 
	\textsuperscript{\rm 1}School of Computer Science, Carnegie Mellon University, Pittsburgh, PA\\
	\textsuperscript{\rm 1}\{byungsoj,namyongp,seojinb\}@cs.cmu.edu\\
}
 
\begin{document}

\maketitle

\begin{abstract}
Massive Open Online Courses (MOOCs) have become popular platforms for online learning.
While MOOCs enable students to study at their own pace, 
this flexibility makes it easy for students to drop out of a class.
In this paper, our goal is to predict if a learner is going to drop out
within the next week, given clickstream data for the current week.
To this end, we present a multi-layer representation learning solution based on branch and bound (BB) algorithm,
which learns from low-level clickstreams in an unsupervised manner,
produces interpretable results, and avoids manual feature engineering.
In experiments on Coursera data, we show that
our model learns a representation that allows a simple model to perform similarly well to more complex, task-specific models, and
how the BB algorithm enables interpretable results.
In our analysis of the observed limitations, we discuss promising future directions.
\end{abstract}

\section{Introduction}
Massive Open Online Courses (MOOCs), such as Coursera, EdX, and Udacity, have become popular online learning methods around the world. MOOCs provide high-quality courses from prestigious faculty, the opportunity for collaborative learning with a global community, and flexibility for students to learn the courses at their own pace \cite{Yang2013TurnOT}. While convenient, this flexibility allows for some students to slow or even completely stop their learning. According to previous studies~\cite{DBLP:conf/lats/YangWHKR15}, between 91\% to 93\% of students dropped or were unable to complete courses. Therefore, if we can predict whether a learner is going to drop out within a week, we can provide appropriate educational treatment for the learners who are most likely to drop out.

How can we predict whether a learner will drop out during the week given a clickstream of learners for the current week? The basic assumption is that the clickstream of a learner connotes certain behavioral patterns of dropout. There are several challenges. 
First, it is hard to extract meaningful behavioral representation of a user from low-level clickstream data, since clickstream data set depends on many different and possibly unknown factors such as user's learning style, syllabus, and content of the week as well as potential noises \cite{SinhaJLD14}. Second, manual feature engineering, on which most existing methods rely, not only requires human efforts, but also is subjective so it might put too much focus on unimportant patterns or miss significant patterns. Third, it is hard to make use of existing approaches to analyze clickstream data. While many previous approaches to predict dropout assume that the input data set arrives at regular intervals, or is of the same length for each observation, clickstream for each learner can be different in terms of both interval and length. To address these challenges, we propose an efficient method to learn meaningful clickstream representation in an unsupervised way so that we can predict dropout based on the representation using simple classification models.

To predict dropout of a learner within the next week, we use low-level clickstreams which convey rich and valuable information, 
but are not easily interpretable and have intrinsic noise. To resolve the problems discussed above, our model (1) summarizes the low-level noisy information into interpretable behavioral, noise-resistant actions by learning a representation of a clickstream using modified Branch and Bound (BB) algorithm and deep neural representation learning using multi-layer perceptron (MLP); (2) avoids hand-tunning of features, which is subjective and may result in significant bias, by selecting the top $k$ ranked features obtained from BB and passing them to an MLP to learn a context-dependent representation of the week given surrounding weeks; and (3) generates a representation for each clickstream applicable to existing deep learning models using BB.

\section{Related Work}
In this section, we illustrate approaches for predicting dropout using clickstream data. We then describe previous methods for learning representations of text based on neural networks.

\textbf{Predicting Dropout Based on Clickstream Data.}
Recently, a considerable number of studies have been conducted to predict whether a user is likely to drop out within the next week based on a wide range of MOOC data sets ~\cite{Sinha2014,DBLP:journals/ce/LykourentzouGNML09,Yang2013TurnOT,kloft2014predicting,nagrecha2017mooc,whitehill2015beyond}. 
Even though the details and types of features they used vary, most studies are based on rich auxiliary features that are not often available, such as demographics, quiz score, forum activities, etc. However, they did not achieve the model to extract rich and valuable information from the clickstream data.

A few studies have used clickstreams to predict dropout, yet most works are based on hand-crafted features extracted from clickstream. \citeauthor{kloft2014predicting} used a linear SVM on deliberately designed features extracted from the clickstream. 
\citeauthor{taylor2014likely} suggested a logistic regression to predict dropout using their predefined features similar to \cite{kloft2014predicting} and additionally the quiz and grade information. 
As they used simple machine learning models such as logistic regression and SVM, they put a lot of effort on hand-crafting on the clickstreams to make it acceptable to the model. Through the hand-crafting of features, valuable information underlying the clickstream might be discarded or 
the model might be damaged by unintentional biases. 

To address that, several deep neural network models have been suggested to utilize clickstreams. 
\citeauthor{fei2015temporal} used deep learning models such as recurrent neural network (RNN) although lightly hand-crafted features are used.
\citeauthor{DBLP:conf/iccse/WangYM17} built deep neural networks using a combination of convolutional neural network archite (CNN) on clickstreams. However, as it simply embeds zero values to make clickstreams of different size be of the same size, unnecessary noises might increase variation of the model. Also, it lacks the interpretability as it uses raw clickstream data. 

To avoid the over/subjective-summarization of clickstream data using hand-engineered features, and under-summarization of clickstream data using raw, noisy clickstreams, we elaborate on building meaningful representation from raw clickstream in an unsupervised manner while minimizing loss of sequential information in clicks.

Several approaches have been suggested to construct cognitively meaningful representations summarized in behavioral levels from raw video clickstream data to predict the dropout \cite{SinhaJLD14,Sinha2014}. 
For instance, \citeauthor{SinhaJLD14} extracted actions by counting the top $M$ most frequent n-grams and further grouped the top actions into predefined behavioral groups. 
Extracting higher-level feature representations from raw clickstream data aims to (1) obtain noise-resistant and interpretable features and (2) transform the unstructured raw clickstream data to structured data that existing statistical or machine learning models can use. However, they still require human efforts to predefine behavioral categories and groups. These works motivated us to build up meaningful representation in algorithmic ways without these human efforts.

\textbf{Neural Representation Learning.}
Recently, neural representation learnings have shown great achievements in a variety of domains~\cite{DBLP:journals/pami/BengioCV13}. 
However, there has not been an approach using neural representation learning to obtain meaningful representation of clickstreams.

An $n$-gram model is a type of probabilistic language model which was originally used for natural language processing \cite{broder1997syntactic}. 
It takes a consecutive sequence of length by moving a window of size $n$ on each sentence and predicts the next item in the sequence. Skip-gram~\cite{DBLP:conf/nips/MikolovSCCD13,DBLP:journals/corr/abs-1301-3781} is a generalization of $n$-grams, 
which has inspired many representation learning methods. 
It does not require the components to be consecutive and can leave gaps that are skipped over. 
%
The intuition behind Skip-gram is that the ``good'' representation of a word represents it in the context surrounding the word. 
Since we want to represent the whole click sequence, not each click, 
previous methods cannot be employed directly to our problem 
as they are designed to learn representations of each word.

There has also been neural representation learning for a variety of sequence. 
\citeauthor{DBLP:conf/nips/KirosZSZUTF15} proposed Skip-thought vectors, a sequence-to-sequence model trained on pairs of consecutive sentences to produce representations for each sentence.
Med2Vec~\cite{DBLP:conf/kdd/ChoiBSCTBTS16} is an approach that learns interpretable representations of both unordered medical codes and ordered visits through multiple layers. 
However, none of these methods are tailored to repetitive and noisy nature of click sequence. We usually have single-digit types of clicks, which produce very repetitive patterns within a sequence. Moreover, a click sequence is noisy as it depends on a variety of variables such as course contents and user's learning styles. 

\section{Proposed Method}
\begin{figure*}[ht]
	\centering
	\includegraphics[width = 0.80\textwidth]{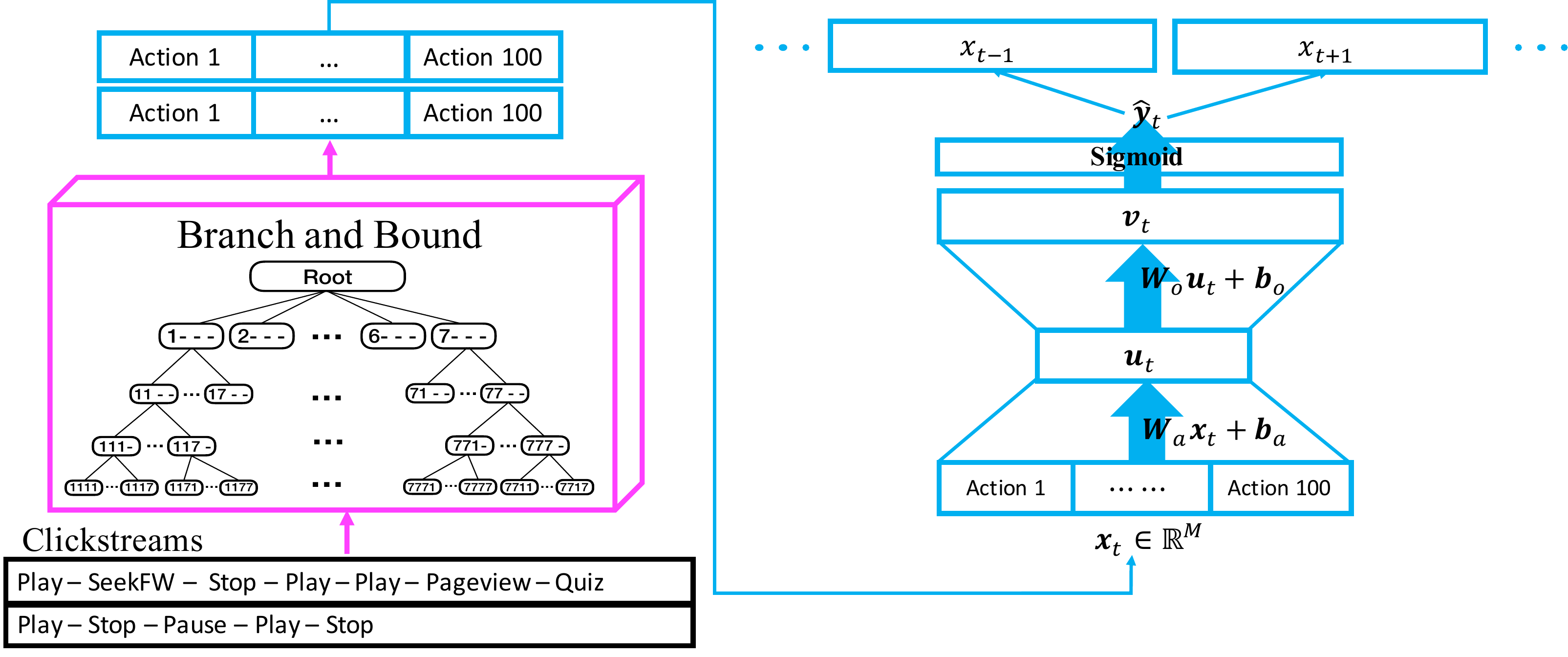}
	\caption{\footnotesize Our proposed model for learning meaningful representations of user actions. Given clickstream data, our modified BB algorithm (left) produces preliminary week-level action representations, which are further refined by our MLP model (right).}
	\label{fig:model}
\end{figure*}

We aim to predict whether a learner will drop out within the next week given a clickstream of learners for the current week. Our specific goal is to build up meaningful and effective representation of a learner's behavioral pattern from the clickstream in an unsupervised manner as in Figure~\ref{fig:model} so that even simple classification models like multi-layer perceptron (MLP) can predict dropout comparably well to more complex task-specific classification models. Given repetitive and noisy clickstreams, our method uses a modified branch and bound algorithm and an MLP to generate interpretable clickstream representations capturing co-occurrence information around the week using Skip-gram method~\cite{DBLP:conf/nips/MikolovSCCD13}. 

\subsection{Branch and Bound}\label{sec:method:bb}
It is hard to understand learners' behavior based on the raw clickstream data that are hardly cognitive. Therefore, we propose the Branch and Bound algorithm that extracts a higher-level representation from the raw clickstream data. The higher-level representation is composed of the scores for top $M$ actions. Here, we define the action as a $n$-gram of clicks that can be closer to the smallest cognitive behavioral unit. For example, `Play' is a click while `Play - Stop - Play' is an action of length 3. The score of action $a$ of length $n$ is
\begin{align}
\sum_{i=1}^{L-n+1} n - d(a,x_i)
\end{align}
where $L$ is the length of a click sequence, $x_i$ is the action of length $n$ starting from $i$-th position in the sequence, and $d(a, x_i)$ is the Hamming distance \cite{hamming1950error} between $a$ and $x_i$, which is the number of clicks that are not matched between two actions of the same length. This action score represents how similar a click sequence $x$ is to the action $a$. 
For example, when $a$ is `Play - Stop - Play' and the click sequence is `Play - Stop - Seek',
the score of action $a$ is $2$. 

To pick top $M$ actions, we rank actions by the standard deviation of the score across week click sequences for each week. Thus, top $M$ actions are more likely to distinguish click sequences than other actions. We do not use the score for all the possible actions because, if so, the dimensionality is $C^n$, which can be huge depending on $n$ and the number of click types, $C$. Once we pick top $M$ actions, we translate a click sequence for each week into the scores of top $M(= 100)$ actions as shown in Figure~\ref{fig:model}.

We use a Branch and Bound mechanism to save the computational cost to pick up top $M$ actions, which motivates the name of our method. The key idea is to search the tree from top to bottom and skips a whole branch if it turns out that all its descendant leaves cannot produce top $M$ that has already been obtained. Thus, it reduces computational cost by not searching all the leaf nodes. The branch-skip decision is simply made by comparing optimal (largest) standard deviation between the highest vertex of the branch and the group of actions with the top $M$ distances. Since a common prefix of actions of its descendants is assigned to the highest vertex, the optimal score would be defined as a standard deviation of the score assuming that the undefined postfix is always matched with the postfix of each $n$-gram of the clickstream.

\subsection{MLP for Learning Feature Representation (MLP-LFR)}\label{sec:method:mlp}
Preliminary action representations obtained from BB contain two types of information: 
the inter-week sequential information and the action-level cooccurrence information.
We use a multi-layer perceptron (MLP) to generate a feature representation that captures both information.

\textbf{MLP Architecture.}
Figure~\ref{fig:model} describes the architecture of an MLP used in our method.
The first MLP layer receives a preliminary action representation $\bx_t \in \R^{M}$ from the BB model, 
and converts it into an intermediate latent representation $\bu_t \in \R^{m}$ via a linear transformation as follows:
\begin{align}
\bu_t = \bW_a \bx_t + \bb_a
\end{align}
where $\bW_a \in \R^{m \times M}$ is the weight matrix for action encoding, and $\bb_a \in \R^{m}$ is the bias vector.
Then, the second MLP layer generates the final action representation $\hat{\by}_t \in \R^{M}$:
\begin{align}
\hat{\by}_t = \text{sigmoid}(\bW_o \bu_t + \bb_o)
\end{align}
where $\bW_o \in \R^{M \times m}$ is the weight matrix for action decoding, and $\bb_o \in \R^{M}$ is the bias vector.
Note that the final action representation $\hat{\by}_t$ we use for predicting dropout is of the same size as the input $\bx_t$.
This facilitates the interpretation of the final dropout prediction 
since each value in the final representation is directly mapped to each action discovered by the BB algorithm.

\textbf{Learning from the Inter-Week Sequential Information.}
A sequence of week-level user actions can be exploited for learning an effective action representation.
We train the MLP to minimize the following error:
\begin{align}
\min_{\bW_{a,o},\bb_{a,o}} \frac{1}{2wT} \sum_{t=1}^{T} \sum_{-w \le i \le w, i \ne 0}^{} \left( \hat{\by}_t - \bx_{t+i} \right)^2
\label{eq:mlp_loss}
\end{align}
where $\hat{\by}_t$ is the final action representation at week $t$, 
$\bx_{t+i}$ is the preliminary action representation at week $t+i$ obtained from the BB algorithm,
$w$ is the context window size, and $T$ is the total number of weeks.
Our intuition for the above loss function is that since learning is a continuous
process for each user, a representation that corresponds to
a user's activities at some point should be able to predict the user's
learning activities in both the recent past and the near future.
Specifically, given a preliminary action representation $\bx_t$ at week $t$,
we consider those action representations $\bx_{t+i}$ that are within a context window 
defined by a tunable parameter $w$, and minimize the mean squared error
between $\hat{\by}_t$ and $\bx_{t+i}$.

\textbf{Learning from the Action-Level Cooccurrence Information.}
Another source of information we can employ is the intra-week cooccurrence information of different actions.
A preliminary action representation $\bx_t \in \R^{M}$ contains scores between 0 and 1 for $M$ actions during week $t$,
from which we can extract representative actions that appeared together in the same period.
Our main idea is that the representation of the actions that occur in the same week should predict each other.
Given a real-valued vector $\bx_t \in \R^{M}$, we define representative actions to be actions whose score is in the top $R$\%, and
consider only representative actions in finding the action cooccurrences.
From a sequence of actions $\bx_1, \bx_2, \ldots, \bx_T$, we maximize the following loglikelihood:
\begin{align}
\begin{aligned}
\min_{\bW_{a}} \frac{1}{T} \sum_{t=1}^{T} \sum_{i:a_i \in \bx_t} \sum_{j:a_j \in \bx_t, j \ne i} \log p(c_j | c_i) \text{~where~}\\
p(c_j | c_i) = \frac{\exp(\bW_{a}[:, j]^{\top} \bW_{a}[:, i])}{\sum_{h=1}^{M} \exp(\bW_{a}[:, h]^{\top} \bW_{a}[:, i]) }.
\end{aligned}
\label{eq:mlp_loss2}
\end{align}

\textbf{Unified Learning.}
Our method learns from both inter-week sequential and intra-week cooccurrence information simultaneously 
from a single source of user actions by combining the above two objective functions as follows:
\begin{align}
\begin{aligned}
\min_{\bW_{a,o},\bb_{a,o}} \frac{1}{T} \sum_{t=1}^{T} \left( - \sum_{i:a_i \in \bx_t} \sum_{j:a_j \in \bx_t, j \ne i} \log p(c_j | c_i) + \right. \\
\left. \frac{1}{2w} \sum_{-w \le i \le w, i \ne 0}^{} \left( \hat{\by}_t - \bx_{t+i} \right)^2 \right).
\end{aligned}
\label{eq:mlp_loss_combined}
\end{align}

\subsection{Dropout Prediction}\label{sec:method:dropout}
We use a simple classification model to predict dropout assuming that our representation trained in an unsupervised way is good enough to work well even with simple models. We use an MLP with three hidden layers whose sizes are 100, 50, 25 and an output layer consisting of a single unit with sigmoid activation, which presents the probability of dropout. Due to the imbalanced classes, margin ranking loss is our loss function for predicting dropout so that we pair one positive instance (dropout) of the user with other negative instance of the same user. We minimize the following objective function:
\begin{align}
\min_{\bW^{l},\bb^{l}} \frac{1}{T} \sum_{t=1}^{T} \max{(0, -(P_{pos} - P_{neg}) + M)}
\end{align}
where $\bW^{l}$ and $\bb^{l}$ are the weights and biases for $l$-th layer from bottom to top given $l = 1,2,3,4$; $P_{pos}$ and $P_{neg}$ are the probability of dropout for positive and negative instances computed from our three-layer MLP; and $M$ is margin, which is set to 0.5 in our experiments.


\section{Experiments}
\subsection{Dataset}\label{sec:exp:dataset}
We use the dataset collected from Coursera\footnote{\url{https://www.coursera.org/}}, the top ranked MOOC platform with more than 28 million users and 2,000 online courses.
This dataset is provided by our partnered faculty \cite{DBLP:conf/lats/YangWHKR15}, and includes 
clickstream data that contain clicks of Coursera learners who took the video lectures of the Microeconomics course for 12 weeks maximum.
It includes 2,709,053 clicks collected from 48,090 users. The clicks are divided into 8 categories: \textit{Pageview}, \textit{Quiz}, \textit{Forum}, \textit{Play}, \textit{Pause}, \textit{Seek}, \textit{RateChg}, and \textit{Stalled}.

\textbf{Preprocessing.}
We encode the 8 click categories to 7 click types: \textit{Pageview, Quiz, Forum, Play, Pause, SeekFW, SeekBW}. \textit{Seek} and \textit{RateChg} are first divided into two categories. By comparing playrates before/after the click event, \textit{RateChg} is divided into \textit{RateChgFast} and \textit{RateChgSlow} and \textit{Seek} is divided into \textit{SeekFW} and \textit{SeekBW}. Next, \textit{RateChgFast} and \textit{RateChgSlow} are merged into \textit{SeekFW} and \textit{SeekBW}, respectively. \textit{Stalled} is removed from the clickstream 
as it is an external issue not related to user's behavioral patterns.
Next, we concatenate the clicks for every user to make them as a continuous clickstream. The clickstream of a user is grouped by a week so that each user can have 1 to 12 clickstreams with different length, called click-level information. The dropout of week $t$ is labeled as 1 if a user dropped out during the next week, i.e. the user's latest clickstream data is the week $t+1$. Clickstreams shorter than the action size $L$ are discarded. After preprocessing, 
the Microeconomics data has 10,904 users having 2.83 weeks per user on average and 30,848 weeks in total. In total, 1,598 weeks are labeled as dropout among 12,104 weeks. 
We named the processed data as \textit{type A}, which is noisy since it includes all noisy week clickstream other than clickstream shorter than $L$. Additionally, we made another data set, called \textit{type B}, less noisy one thanks to additional preprocessing 
where we removed clickstreams composed of 1,000 or more clicks, and users dropped before the fourth week. After preprocessing, 
\textit{type B} data has 1,598 users having 7.57 weeks per user on average, and 12,104 weeks in total. 

\subsection{Experimental Settings}\label{sec:exp:settings}
We implement the BB algorithm (Section~\ref{sec:method:bb}) for action search. We set the action size $L$ to 4, and select the top $M$ actions ($M = 100$) with the largest standard deviation of the average Hamming distance. With this setting, the BB algorithm reduces the clickstream data to structured data in which each sample consists of the same 100 actions.
We implement our MLP model for feature representation learning (Section~\ref{sec:method:mlp}) in PyTorch.
Based on the results obtained with several parameter values, 
we train our MLP model with the stochastic gradient descent (SGD) algorithm using the following settings: input dimension $M$ $(100)$, hidden layer size $m$ $(20)$, 
context window size $w~(1)$, learning rate $(0.01)$, momentum $(0.9)$, weight decay $(0.0001)$, and batch size $(200)$. We train our MLP model until the maximum number of epochs (1000) has been reached, or the difference of the loss over a validation set (15\% of the data) between consecutive epochs gets less than a threshold (0.000001).
Our dropout prediction layer (Section~\ref{sec:method:dropout}) is trained with the Adam optimizer ($\beta_1=0.9$ and $\beta_2=0.999$) until convergence using the following parameter settings: learning rate (0.001), batch size (10), momentum $(0)$, weight decay $(0)$, and margin $M$~$(0.5)$.

\subsection{Quantitative Results}\label{sec:exp:results}
In order to evaluate our approach, we compare our model and our model without MLP-LFR (Section~\ref{sec:method:mlp}) with two fairly complex baseline approaches: 1D-CNN~\cite{hu2014convolutional} and 1D-CRNN~\cite{DBLP:conf/iccse/WangYM17}. Here, RNN-like method is excluded because week clickstream is too long to train RNN. Baseline 1D-CNN model takes a clickstream of a week as an input and consists of embedding layer whose size is 20, two convolutional layer whose filter size and number of output are 3 and 32, two max pooling layer whose stride is 3, and one softmax output layer. 1D-CRNN is same as in~\cite{DBLP:conf/iccse/WangYM17}. Our model without MLP-LFR uses the same dropout prediction model with the BB algorithm but without MLP-LFR, which aims to show the contribution of the MLP-LFR in our model. We measure the performance based on F1 score and AUC due to the imbalance of positive and negative instances in our dataset. Also, we tested on the type A and B datasets to see how the performance changed on different kinds of datasets. 

\begin{table}[ht]
	\centering
	\caption[caption]{{\bf Model evaluation on type A (top) and type B (bottom) datasets in terms of F1 score and AUC.}}
	\label{table:modeleval}
	\begin{adjustbox}{max width=\textwidth}
		\begin{threeparttable}
			\setlength{\tabcolsep}{3.0pt}
			\begin{tabular}{@{}lcccc@{}}
				\toprule
				~ & 1D-CNN & 1D-CRNN & \makecell{Our Model\\ w/o MLP-LFR} & Our Model\cr
				\midrule
				F1 score & 0.4252 & N/A & \textbf{0.4364} & 0.4124\\
				AUC & 0.6023 & N/A & \textbf{0.6131} & 0.5942\\
				\bottomrule
			\end{tabular}
			\begin{tablenotes}
			  \tiny
			\item
			\end{tablenotes}
		\end{threeparttable}
	\end{adjustbox}
	\begin{adjustbox}{max width=\textwidth}
		\begin{threeparttable}
			\setlength{\tabcolsep}{3.0pt}
			\begin{tabular}{@{}lcccc@{}}
				\toprule
				~ & 1D-CNN & 1D-CRNN & \makecell{Our Model\\ w/o MLP-LFR} & Our Model\cr
				\midrule
				F1 score & \textbf{0.3809} & 0.3728 & 0.3747 & 0.3428 \\
				AUC & 0.7290 & 0.7321 & \textbf{0.7385} & 0.6743 \\
				\bottomrule
			\end{tabular}
		\end{threeparttable}
	\end{adjustbox}
\end{table}


%
Table~\ref{table:modeleval} show the results on the noisy, type A dataset (top) and 
the type B dataset (bottom), which is cleaner than type A dataset thanks to additional preprocessing. We cannot run 1D-CRNN on type A dataset since type A dataset includes very long clickstream whose length is more than ten thousand, which cannot be handled by 1D-CRNN. On both datasets, we can see that our model without MLP-LFR, which includes only the BB algorithm part and is trained in an unsupervised manner, achieves comparable performance to fairly complex, task-specific baseline models in terms of AUC and F1 score. Note that we only use a simple MLP for dropout prediction, that does not take the sequential nature of data into account unlike our baselines. The result indicates that the learned representation from the BB algorithm captures signals that are as informative as those from uninterpretable baselines while maintaining the interpretability. Yet, our model with MLP-LFR fails to achieve comparable performance on type A and B datasets. Shortly, our MLP-LFR does not work well as we intended while our BB algorithm works pretty well. We give a discussion on these results in Section~\ref{sec:discussion}.

\subsection{Qualitative Results}\label{sec:exp:qual_results}
We qualitatively characterize weeks that a learner drops out and that a learner does not drop out by taking advantage of interpretable action scores computed by the BB algorithm. We identified the actions that show significantly different average action scores between the non-dropout and dropout weeks. For each week, we computed the score for each action by the BB algorithm and compared its mean between the two groups (non-dropout and dropout weeks) using the two-sample t-test. Table~\ref{table:toptenactions} shows the top 10 actions that characterize the non-dropout (top) and dropout (bottom) groups: the top 10 actions having the smallest t-score characterize the non-dropout group while the top actions having the largest t-score characterize the dropout group. The non-dropout group are characterized by the actions including ``Quiz'', which means successful learners are likely to take quizzes that may intrigue their interest and help to stay motivated. Meanwhile, the dropout group is characterized by the actions including ``SeekBw'', ``Pause'', and ``SeekFw'' which can be interpreted that the learners are likely to drop out when they struggle with difficult concepts.
\begin{table}[ht]
	\centering
	\caption[caption]{{\bf The top 10 actions of non-dropout (top) and dropout (bottom) weeks}}
	\label{table:toptenactions}
	\begin{adjustbox}{max width=\linewidth}
		\begin{threeparttable}
			\begin{tabular}{@{}ccc@{}}
				\toprule
				Action & t-score & p-value \cr
				\midrule
				SeekBw SeekFw SeekBw Quiz & -7.195&$<$ 2e-16\cr
				SeekFw SeekBw SeekBw Quiz & -7.188&$<$ 2e-16\cr
				SeekBw SeekBw SeekFw Quiz & -7.176&$<$ 2e-16\cr
				SeekBw SeekFw Quiz SeekBw & -6.772&2.64e-15\cr  
				SeekFw SeekBw Quiz SeekBw & -6.763&2.81e-15\cr
				SeekBw SeekBw Quiz SeekFw & -6.760&2.87e-15\cr
				SeekFw Quiz SeekBw SeekBw & -6.676&5.10e-15\cr 
				SeekBw Quiz SeekBw SeekFw & -6.673&5.20e-15\cr 
				SeekBw Quiz SeekFw SeekBw & -6.668&5.40e-15\cr 
				SeekBw SeekBw SeekBw Quiz & -6.625&7.21e-15\cr
				\bottomrule
			\end{tabular}
			\begin{tablenotes}
			  \tiny
			\item
			\end{tablenotes}
		\end{threeparttable}
	\end{adjustbox}
	\begin{adjustbox}{max width=\linewidth}
		\begin{threeparttable}
			\begin{tabular}{@{}ccc@{}}
				\toprule
				Action & t-score & p-value\cr
				\midrule
				SeekFw SeekFw SeekFw SeekFw & 1.635 &0.000395\cr
				SeekBw SeekFw SeekFw SeekFw & 1.232 &0.000371\cr
				SeekFw SeekBw SeekFw SeekFw & 1.102 &0.000360\cr
				SeekFw SeekFw SeekBw SeekFw & 1.080 &0.000358\cr
				Pause SeekFw SeekFw SeekFw  & 1.047 &0.000355\cr
				SeekFw SeekFw SeekFw SeekBw & 1.020 &0.000352\cr
				SeekFw Pause SeekFw SeekFw & 0.911 &0.000341\cr
				SeekBw SeekBw SeekBw SeekBw & 0.838 &0.000333\cr
				Pause SeekBw SeekBw SeekBw & 0.811 &0.000330\cr
				SeekFw SeekFw Pause SeekFw & 0.810 &0.000330\cr
				\bottomrule
			\end{tabular}
		\end{threeparttable}
	\end{adjustbox}
\end{table}

\section{Discussion and Conclusion}\label{sec:discussion}

In this section, we give a summary of our analysis on the results, and provide a detailed discussion on the limitations of our method, and 
conclude with a plan for future works.

Our MLP-LFR does not work as well as we intended while our BB algorithm achieves comparable performance to fairly complex task-specific baselines even though it is trained in an unsupervised fashion. We hypothesize that our MLP-LFR fails to achieve the improvement either because our BB algorithm loses useful sequential information within a week at the expense of the interpretability, or because ``a week as a unit'' is an inappropriate strategy to divide a clickstream. We could have used more than one week's clickstreams, but we did not because our focus was on exploiting sequential information over weeks. Another potential reason why our MLP-LFR does not show much improvement is that weekly clickstreams in our dataset have weeks in which no click is observed, which may lead to the violation of our assumption on the temporal dependency between consecutive weeks. 

In retrospect, we identify three limitations of our work.
First, as briefly discussed, MLP-LFR fails to learn the representation that encodes meaningful sequential information. Considering that the course content is one of the most influential factors on users' learning and users' click patterns, we conjecture that this is primarily because MLP-LFR does not consider a student's progress in the lecture. Instead, our model takes a rather straightforward approach of working with a sequence of week-level action representations. Since students progress through courses at a different pace depending on the course content and their current level of understanding, one student may take a much longer time than other students in completing the course, and as a result, the variance of week clickstream  may be very high, hindering the training of MLP-LFR. Therefore, we assume that it would be more meaningful and effective to learn action representations based on the actual learning progress of each student.
Second, our model is designed to perform best when given consecutive clickstream data since we learned a context-dependent representation of a week in surrounding weeks. However, preprocessed clickstreams may not be consecutive in terms of weeks. This is either because some weeks may get filtered out during the preprocessing step, 
or because there can be gaps in the input clickstream in case users took courses infrequently, e.g., taking a course once every few weeks.
Third, our model does not use personal information (e.g., the age and highest education level of each student), or auxiliary week-level
information (e.g., clickstream length), which can provide nontrivial
information about user behavior.

We plan to address the above limitations on four different aspects. 
First, we will preprocess clickstream data differently such that action representations are generated based on users' actual learning progress instead of a time sequence. For example, clickstreams would be divided into several sub-sequences that occur in the same lecture video instead of the same week. This may be a more meaningful unit than a week since users' behavioral pattern may be more consistent within a lecture than within a week.
Second, we plan to modify the loss function of the representation learning to resolve nonconsecutive interval between weeks/sessions. We can introduce a decay parameter for the weight on surrounding action representations, which decreases as the time interval between weeks/sessions increases.
Third, we plan to incorporate auxiliary information such as weekly quiz score and demographic information into the model. One possible approach would be to append the additional information to the intermediate action-level representation. For example, the week-level information would be merged to the intermediate layer in MLP-LFR. 
%
Fourth, we also plan to improve our simple dropout prediction model by considering sequential information.
For example, we can generate more fine-grained sequence by dividing a week into several sessions of shorter length. 
%


{
\fontsize{10.0pt}{11.0pt} \selectfont
\bibliography{reference}
\bibliographystyle{aaai}
}
\end{document}